# Ground-SLAM: Ground Constrained LiDAR SLAM for Structured Multi-Floor Environments

Xin Wei[1], Jixin Lv[1], Jie Sun, Shiliang Pu*

*Abstract*—This paper proposes a 3D LiDAR SLAM algorithm named Ground-SLAM, which exploits grounds in structured multi-floor environments to compress the pose drift mainly caused by LiDAR measurement bias. Ground-SLAM is developed based on the well-known pose graph optimization framework. In the front-end, motion estimation is conducted using LiDAR Odometry (LO) with a novel sensor-centric sliding map introduced, which is maintained by filtering out expired features based on the model of error propagation. At each key-frame, the sliding map is recorded as a local map. The ground nearby is extracted and modelled as an infinite planar landmark in the form of Closest Point (CP) parameterization. Then, ground planes observed at different key-frames are associated, and the ground constraints are fused into the pose graph optimization framework to compress the pose drift of LO. Finally, loop-closure detection is carried out, and the residual error is jointly minimized, which could lead to a globally consistent map. Experimental results demonstrate superior performances in the accuracy of the proposed approach.

## I. INTRODUCTION

Simultaneous Localization And Mapping (SLAM) plays a critical role in various applications, such as service robots, patrol robots and autonomous driving. SLAM is challenging since it is a "chicken-egg" problem. Extensive efforts have been made on improving the accuracy and robustness using different kinds of sensors such as camera, Inertial Measurement Units (IMU), Global Positioning System (GPS) and Light Detection And Ranging (LiDAR) [1]. LiDAR provides accurate measurement of the surroundings in the form of point cloud. The measurement is robust to variations of lighting conditions [2], making LiDAR popular in the SLAM field, especially for outdoor applications such as autonomous driving.

Typical 3D LiDAR SLAM algorithms utilize the pose graph optimization framework [3] consisting of front-end and back-end. In the front-end, successive LiDAR scans are aligned using point cloud registration algorithms, and the sensor trajectory is recursively estimated. This process is commonly referred as LiDAR Odometry (LO) [4]. Iterative Closest Point (ICP) [5], [6] and Normal Distribution Transformation (NDT) [7] are two widely applied algorithms for point cloud registration. For mapping applications that require high accuracy, scan-to-map method [4] is frequently adopted instead of the scan-to-scan approach. In the back-end, loop closure detection is conducted to recognize the re-visited

X. Wei, J. Lv, J. Sun, S. Pu are with Hikvision Research Institute (HRI), Hangzhou Hikvision Digital Technology Co.,Ltd., Hangzhou, China. Email: {weixin7, lvjixin, sunjie, pushiliang.hri}@hikvision.com.
[1]Both authors contributed equally to this manuscript.
*Corresponding author.

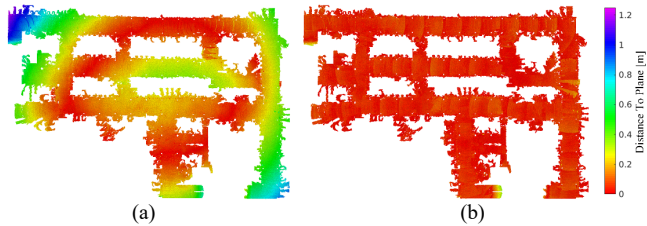

Figure 1. Top view of distances between ground points of mapping results and fitted ground plane. (a) Predictable pose drifts happens due to LiDAR measurement bias. To be specific, SLAM result drifts upward when the mapping vehicle is moving on the ground plane. As a result, the assembled ground is distorted as an upward bowl. (b) The assembled ground becomes flat after ground observation constraints are fused into SLAM process.

place [8]. And a pose graph is constructed, the residual error is jointly minimized using non-linear least square optimization algorithms [3], [9].

Extensive research has been done to improve the accuracy and robustness [10] of 3D LiDAR SLAM results, such as multi-sensor fusion [11], [12], cooperating promising end-to-end techniques [13] and semantic information [14], [15] to SLAM. As far as our best concern, most LO approaches maintain the sliding map in their scan-to-map process using the range-based method. This kind of method roughly removes the map's features that are out of a pre-set range from the current state [4]. It is inefficient for redundant feature removal because a lot of unobservable features within the range are kept. Besides, it could bring inconsistency into the maintained map in specific scenarios. This situation occurs when LiDAR moves around small buildings where the observation is partially occluded, and the pose error accumulates in the limited area, as presented in Fig. 2. In addition, LiDAR measurement noise and its impact on SLAM results haven't gained enough attention in the robotics community. Recently, J. Laconte [16] finds that LiDAR measurement bias exists and could reach up to 20 cm for high incidence angles. This situation happens when LiDAR scans on the surface of the road in the far distance. As a consequence, the points observed from the road are slightly bent, and the trajectory estimated by LO is prone to drift along the vertical direction of the road, as shown in Fig. 1(a). This drift error can be easily compressed in the outdoor application by fusing global pose observation such as Real-Time Kinematic (RTK)-GPS into the optimization process. However, global pose observation becomes difficult to obtain in the indoor scenario and the SLAM result is prone to be distorted.

In this paper, we propose a 3D LiDAR SLAM algorithm that exploits ground constraints in structured multi-floor environments to improve the accuracy of SLAM results. The key contributions of this paper are as follows:

- A sensor-centric sliding map maintenance method, which can efficiently eliminate redundant features and keep essential features using an observation-based method, is proposed.
- A local ground correspondence method, which associates ground planar landmarks between consecutive key-frames, is proposed.
- A 3D LiDAR SLAM algorithm, which exploits grounds in indoor multi-floor environments to compress drift caused by LiDAR measurement bias, is proposed.

The rest of this paper is organized as follows: Sec. II presents a review of related work. Some notations and preliminaries are explained in Sec. III. Sec. IV presents a detailed explanation of the Ground-SLAM system. And the qualitative and quantitative tests and analyses are demonstrated in Sec. V. Sec. VI presents the conclusion along with some future work.

## II. RELATED WORK

Visual and LiDAR SLAM are popular topics in the robotics community and have been extensively investigated during the past two decades. This paper focuses on approaches based on 3D LiDAR, together with some algorithms exploiting ground constraints. For a more detailed review of recent SLAM work, please refer to article [17] and the references therein.

### A. 3D LiDAR SLAM

LO is a fundamental component of LiDAR SLAM. Though the measurements of LiDAR are mostly accurate and robust, some vital shortcomings of LiDAR make the task of building an accurate and robust LO difficult.

First, most mobile LiDAR has a low vertical resolution, and the sparse point cloud it obtains makes robust feature detection and tracking difficult. Therefore, a local map strategy that aggregates past observations and poses is usually utilized in LO [4], [8], [14]. Unfortunately, the dense local map makes point cloud registration time-consuming. From this point, LOAM [4] is a milestone in terms of efficiency and has inspired amount of other work [19], [20]. LOAM extracts and aligns sparse corner and plane features in the odometry process and periodically aligns dense features in the mapping process to balance efficiency against accuracy. SuMa [21] is also impressive, which utilizes a projection method to avoid explicitly finding the closest point. Suma also estimates normal vectors rapidly using the cross product of neighbor vertices. IMLS-SLAM [22] actively samples those points that could provide constraints for pose estimation. With this sampling strategy, only about 10% of the raw LiDAR points would be sampled.

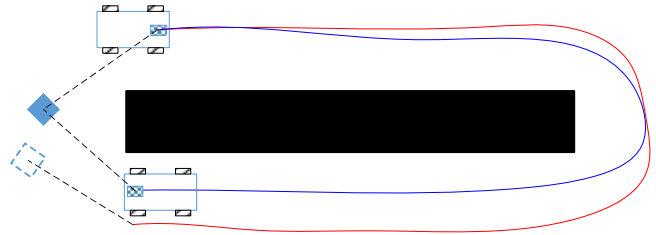

Figure 2. The blue line stands for true trajectory of the LiDAR while the red line represents the trajectory estimated by LO. Due to the inevitable pose drift, the feature observed at first pose is not coincident with the one observed at last pose.

The second disadvantage of LiDAR measurement is that it is not a snapshot of environments and thus suffers from motion distortion. This hazardous effect has been gradually realized for the past few years. Typical motion compensation or de-skewing approaches integrate IMU measurements to de-skew raw LiDAR points [4], [11], [18], [23]. A constant velocity model is usually utilized [4] when the IMU measurements are not available. In [18], the de-skewing process is fused into graph optimization by separating raw scans into several scan lines, and interpolates the corresponding poses using continuous cubic B-spline in SE3.

Furthermore, LiDAR measurement's noise has been modelled as zero-mean Gaussian distribution for simplicity [24] during the past two decades. The influence of incidence angle on the measurement noise is conducted in [25], while bias has not been considered. Recently, J. Laconte [16] finds that LiDAR measurement bias exists especially for high incidence angle and would lead to predictable drifts. Besides, a physical explanation and model of this bias is presented, and it is used to remove this bias from LiDAR measurements. Nevertheless, we argue that the incidence angle is difficult to estimate during the SLAM process. Besides, LiDAR measurement bias might also be affected by other factors. Thus, instead of estimating bias by introducing the model as in [16], we exploit ground constraints to compress drifts and the mapping error caused by LiDAR measurement bias in indoor multi-floor environments.

### B. SLAM with Ground Constraints

The ground constraint is widely utilized in various SLAM systems. F. Zheng [26], [27] proposes a method for ground vehicles that fuses ground constraints into the optimization framework. The vehicle pose is constrained by virtual SE2 ground constraints in [26] or directly parameterized by SE2 parameterization [27] and couldn't handle multi-floor environments. Koide [28] assumes that one global ground plane exists, which doesn't hold in multi-floor scenes. Ground points are also extracted and aligned to estimate 3 DOF of 6 DOF pose in LeGO-LOAM [19] and [34]. However, they haven't fused ground information into the pose graph optimization framework. In contrast to these approaches, our method introduces SE3-Plane ground constraints into the optimization frame to improve SLAM results' accuracy.

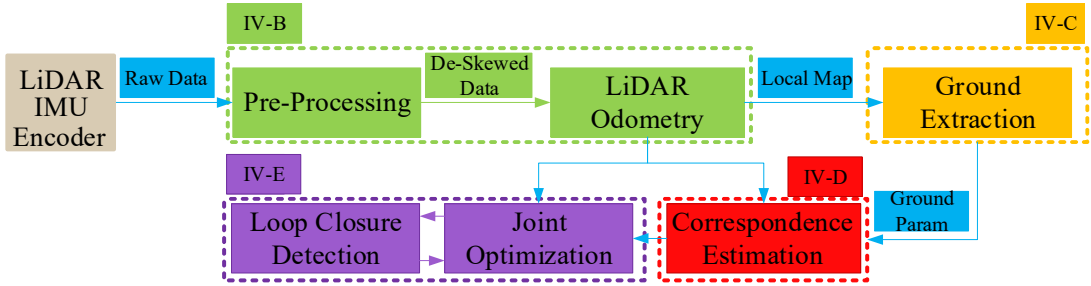

Figure 3. A brief overview of Ground-SLAM framework.

## III. NOTATIONS AND PRELIMINARIES

### A. Notations

In the following section, we denote the homogeneous transformation matrix as $T_a^b \in SE(3)$, which transforms a point $\boldsymbol{p}_i^a \in R^3$ in frame $\mathbf{F}_a$ into frame $\mathbf{F}_b$. $R_a^b \in SO(3)$ and $\boldsymbol{t}_a^b \in R^3$ is the rotation matrix and translation vector of $T_a^b$, respectively. We use L, B, W to represent the coordinate frame of LiDAR, IMU, and global map, respectively. We use $\mathbf{\Pi}^a$ to represent the Closest Point (CP) parameter of an infinite plane in frame $\mathbf{F}_a$.

### B. Preliminaries

Researchers have proposed several parameterization methods of the infinite plane. Hesse Form (HF) is composed of a normal vector $\vec{n}$ and distance $d$ between the plane and the origin of a given coordinate frame. HF is an over-parameterization since it adopts 3D vectors to represent normal vectors with 2 DOF. Consequently, it suffers from singular information matrix in least-squares optimization, and the identity of the normal vector becomes hard to hold. Spherical coordinate, which represents a normal vector with azimuth and elevation angle, is a minimal parameterization. However, it suffers from ambiguities when the elevation angle is equal to $\pm\pi/2$. Unit quaternion [30] is also exploited to represent the infinite plane. However, the physical connection between the quaternion and plane is unclear.

Inspired by [31], CP, which is defined by the closest point on the plane to the origin of a given coordinate frame, is adopted to represent the infinite plane in this paper. CP is a minimal representation, thus also suffers from the singularity problem when the origin of a given coordinate frame lies on the plane. However, the singularity of CP representation could be easily avoided in this paper since the ground plane has a certain distance from LiDAR equipped on vehicles.

Transforming the ground's parameter between different frames is essential as the ground is modelled as an infinite planar landmark in this paper. HF makes it easy to represent plane equation, thus we will make use of HF as an intermediate to transform ground parameters. Supposing there is a plane with HF $[\vec{n}^a, d^a]$ in frame $\mathbf{F}_a$. Point $\boldsymbol{p}_i^a$ lies on this plane, then this point satisfies the plane equation, as shown in (1).

$$(\vec{n}^a)^T \boldsymbol{p}_i^a = d^a \quad (1)$$

If the transformation matrix $T_b^a$ between frame $\mathbf{F}_a$ and frame $\mathbf{F}_b$ is known, then we have $\boldsymbol{p}_i^a = R_b^a \boldsymbol{p}_i^b + \boldsymbol{t}_b^a$. Substituting this into (1), and we could get (2).

$$((R_b^a)^T \vec{n}^a)^T \boldsymbol{p}_i^b = d^a - (\vec{n}^a)^T \boldsymbol{t}_b^a \quad (2)$$

Then we could get $\vec{n}^b$ and $d^b$ by using equations in (3). After getting the transformation result that is represented in HF, we could transform it into CP parameterization using (4).

$$\begin{cases} \vec{n}^b = (R_b^a)^T \vec{n}^a \\ d^b = d^a - (\vec{n}^a)^T \boldsymbol{t}_b^a \end{cases} \quad (3)$$

$$\mathbf{\Pi}^b = d^b \vec{n}^b \quad (4)$$

## IV. GROUND CONSTRAINED LiDAR SLAM

### A. System Review

A brief review of the proposed framework is shown in Fig. 3. In the pre-processing module, IMU and wheel encoder measurements are fused using the Extended Kalman Filter (EKF) methods to provide high-frequency motion estimation result. The motion distortion of the LiDAR scan is diminished by using the motion estimation result.

The de-skewed LiDAR scan points are used to estimate the relative transformation between consecutive scans using the point-to-plane ICP algorithm. A sensor-centric sliding map is maintained to overcome the sparseness of LiDAR scan and improve the accuracy of LO. The sensor-centric sliding map is recorded as a local map at each key-frame, and the ground plane is extracted using weighted least-square methods. Local correspondences between ground planes observed at different key-frames are determined.

Then ground observation constraints are fused to the pose graph optimization framework. The pose drift, especially the vertical drift mainly caused by LiDAR measurement bias, is compressed. And the accuracy of LO's result trajectory is improved. Afterwards, the loop-closure detection is carried out, and the loop-closure edges between newly associated key-frames are added to the pose graph. Finally, the residual error is minimized, and a globally consistent map is assembled. The details of the critical modules are introduced below.

### B. LiDAR Odometry with Sensor-Centric Sliding Map

Our LiDAR Odometry approach mainly consists of two parts: a scan-to-map registration framework and an observation-based maintenance method for the sensor-centric sliding map.

The primary functions of the registration framework are scan registration and data transformation. Assuming at time $k$, the global pose of LiDAR is $T_{L_k}^W$ and the sliding map $M_{L_k}$ has been maintained w.r.t. frame $L_k$. When a new scan $S_{k+1}$ arrives at time $k+1$, the sensor transformation $T_k^{k+1}$ as well as the corresponding covariance $\Sigma_{T_k^{k+1}}$, are firstly estimated by using point-to-plane ICP [32] and Censi's method [33], where $T_k^{k+1}$ is the abbreviation of $T_{L_k}^{L_{k+1}}$. The coordinate and uncertainty covariance of points in the sliding map is transformed from frame $L_k$ to frame $L_{k+1}$ using (5) and (6), where $\boldsymbol{p}_i^k$ is $i$-th point in the sliding map $M_{L_k}$ and $\Sigma_{\boldsymbol{p}_i^k}$ is the corresponding covariance. $\Sigma_{R_k^{k+1}}$, $\Sigma_{t_k^{k+1}}$ are the covariance matrix of estimated rotation and translation component, respectively. $J_{R_k^{k+1}}$, $J_{\boldsymbol{p}_i^k}$ are the jacobian matrix w.r.t. rotation component and the point, respectively.

$$\boldsymbol{p}_i^{k+1} = T_k^{k+1} \cdot \boldsymbol{p}_i^k = R_k^{k+1} \cdot \boldsymbol{p}_i^k + \boldsymbol{t}_k^{k+1} \quad (5)$$

$$\Sigma_{\boldsymbol{p}_i^{k+1}} = J_{R_k^{k+1}}^T \Sigma_{R_k^{k+1}} J_{R_k^{k+1}} + J_{\boldsymbol{p}_i^k}^T \Sigma_{\boldsymbol{p}_i^k} J_{\boldsymbol{p}_i^k} + \Sigma_{t_k^{k+1}} \quad (6)$$

The observation-based maintenance method, which is designed for the sensor-centric sliding map updating, contains the following steps:

1) First, the registered new scan points are associated with the sliding map points by checking pre-defined distance metric, such as Euclidean distance, Mahalanobis distance, etc.

2) Then, the associated sliding map points are selected, and their uncertainties are reset to their associated new scan points' covariance matrixes, which is commonly referred to as observing error.

3) The sliding map points that capture considerable uncertainty will be filtered. In practice, the point is eliminated as long as the corresponding covariance matrix's trace is more significant than the threshold.

4) Finally, the registered scan points that are failed to be associated are added to the sliding map as new observations.

The observation-based maintenance method filter out the points with considerable uncertainty while the observing features are kept and updated. Therefore, the consistency and density of the sliding map have been preserved, which is the vital requirement for the following scan registration and motion estimation.

*C. Ground's Extraction*

To utilize geometrical ground constraints, we introduce a sound assumption that supposes the ground points near the vehicle could be modelled as an infinite plane. The ground points close to the vehicle's moving path are firstly segmented from the sensor-centric sliding map at each key-frame. Then, an initial infinite plane CP parameter $\boldsymbol{\Pi}_0$ is estimated by applying the RANdom SAmple Consensus (RANSAC) algorithm on the segmented points.

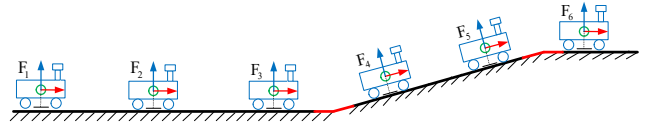

Figure 4. Frames and transformations during ground correspondence estimation.

The initial plane parameter estimated by RANSAC might be noisy. Therefore, we formulate a weighted least squares optimization problem as in (7) and (8), and minimize the point-to-plane distances between ground points and the CP parameterized ground plane.

$$\boldsymbol{\Pi} = \text{argmin} \sum_{i=1}^N \boldsymbol{r}_i^T \Omega_{r_i} \boldsymbol{r}_i \quad (7)$$

$$r_i = \frac{\boldsymbol{p}_i^T \boldsymbol{\Pi}}{\|\boldsymbol{\Pi}\|} - \|\boldsymbol{\Pi}\| \quad (8)$$

Where $\boldsymbol{p}_i$ is the $i$-th point of ground points, $N$ is the number of ground points, $\Omega_{r_i}$ is the inverse of the uncertainty of the $i$-th measurement $r_i$, as in (9), where $\Sigma_{\boldsymbol{p}_i}$ is the uncertainty of measurement $\boldsymbol{p}_i$ and $\boldsymbol{\Pi}_l$ is the linearization point. The cost in (7) is minimized using the Gauss-Newton method. The Jacobian of residual $r_i$ at $\boldsymbol{\Pi}_l$ is formulated as in (10). Then the local increment could be formulated in (11). And the plane parameter could be updated in (12). The corresponding uncertainty is approximated as in (13).

$$\Omega_{r_i} = \left(\frac{\boldsymbol{\Pi}_l^T \Sigma_{\boldsymbol{p}_i} \boldsymbol{\Pi}_l}{\|\boldsymbol{\Pi}_l\|_2^2}\right)^{-1} \quad (9)$$

$$J_i = \frac{\boldsymbol{p}_i^T}{\|\boldsymbol{\Pi}_l\|} - \frac{\boldsymbol{\Pi}_l^T}{\|\boldsymbol{\Pi}_l\|} - (\boldsymbol{p}_i^T \boldsymbol{\Pi}_l)\frac{\boldsymbol{\Pi}_l^T}{\|\boldsymbol{\Pi}_l\|^3} \quad (10)$$

$$\Delta \boldsymbol{\Pi} = -\left(\sum_{i=1}^N J_i^T \Omega_{r_i} J_i\right)^{-1}\left(r_i \Omega_{r_i} J_i\right) \quad (11)$$

$$\boldsymbol{\Pi} = \boldsymbol{\Pi}_l + \Delta \boldsymbol{\Pi} \quad (12)$$

$$\Sigma_{\boldsymbol{\Pi}} = \left(\sum_{i=1}^N J_i^T \Omega_{r_i} J_i\right)^{-1} \quad (13)$$

*D. Ground Correspondence Estimation*

Since the ground is modelled as a planar landmark in this paper, ground planes that are extracted at different positions need to be associated. Fig. 4 shows a classical scenario for ground correspondence estimation in structured multi-floor indoor environments. Supposing there are several sequential key-frames $\boldsymbol{F}_i$, where $i = 1, ..., N$, $N$ is the number of key-frames. Each key-frame possesses an estimated pose $T_i^w$ provided by LO, and an observed ground plane with the estimated parameter $\boldsymbol{\Pi}^i$ and its uncertainty $\Sigma_{\boldsymbol{\Pi}^i}$, then the remaining problem is to determine correspondences between these observed ground planes.

A straightforward solution to this problem is associating ground planes by comparing their parameters in one identical coordinate frame under distance metrics (e.g., Euclidean). However, LO's error accumulates, and this error will be propagated into plane parameters. Consequently, the wrong data association might occur. It should be noticed that LO drifts slightly and remains high precision for a short-term period. Besides, the indoor ground is usually well structured and the ground parameter changes sharply in multi-floor junctions, as shown in Fig. 4. Therefore, we conduct local ground correspondence estimation between consecutive two key-frames by detecting sharp changes of ground's CP parameter, as in (14) and (15), function $f$ refers to the transformation in (3) and (4), and $\Omega_{\Delta\Pi^i}$ is the inverse of the uncertainty of the relative measurement $\Delta\Pi^i$.

$$\Delta\Pi^i = \Pi^i - f(\Pi^{i+1}, T_i^{i+1}) \quad (14)$$

$$e(\Delta\Pi^i) = (\Delta\Pi^i)^T \Omega_{\Delta\Pi^i} \Delta\Pi^i \quad (15)$$

Uncertainty of variables $\Pi^i$, $\Pi^{i+1}$ and $T_i^{i+1}$ in (14) will be propagated to $\Delta\Pi^i$. Accurate geometrical constraints provided by LiDAR points make LO remain high-fidelity for a short-term period. Thus the uncertainty of $T_i^{i+1}$ is small enough to be neglected. Then we can estimate the uncertainty of relative observation $\Delta\Pi^i$ as in (16) and (17), where $R$ and $t$ is the rotation matrix and translation vector of $T_i^{i+1}$, respectively.

$$\left(\Omega_{\Delta\Pi^i}\right)^{-1} = \underbrace{\left(\frac{\partial \Delta\Pi^i}{\partial \Pi^i}\right)^T}_{I_{3\times3}} \Sigma_{\Pi^i} \frac{\partial \Delta\Pi^i}{\partial \Pi^i} + \left(\frac{\partial f}{\partial \Pi^{i+1}}\right)^T \Sigma_{\Pi^{i+1}} \frac{\partial f}{\partial \Pi^{i+1}} \quad (16)$$

$$\frac{\partial f}{\partial \Pi^{i+1}} = R^T - \frac{R^T \Pi t^T + \Pi^T t R^T}{\|\Pi\|_2^2} + \frac{2\Pi^T t R^T \Pi\Pi^T}{\|\Pi\|_2^4} \quad (17)$$

*E. Joint Pose Graph Optimization*

After correspondences between observed ground planes have been determined, ground observation constraints are fused into the pose graph optimization framework to compress the pose drift of LO, as shown in Fig. 5(b). Assuming there is a ground plane **P** whose CP parameter in global coordinate frame W is denoted as $\Pi^w$, frame $F_j$ with pose $T_j^w$ observes **P** with the CP parameter $\Pi^j$, which could be transformed to HF [$\vec{n}^j, d^j$]. With (3) and (4), we could transform the ground plane observation in frame $F_j$ into global coordinates. And we could use the difference between the observed and estimated CP parameters as the residual error, as shown in (19), where $R_j^w$ and $t_j^w$ is the rotation matrix and translation vector of $T_j^w$ respectively. The Jacobian matrix of residual error $r_{ij}$ with respect to $\Pi^w$ and $T_j^w$ is shown in (20), (21) and (22). Then the residual errors of ground constraints and LO are jointly minimized using Levenberg-Marquardt (LM) algorithm, as shown in (18), where $C$ means constraints sets, $r_{ij}$ is an abbreviation of $r(x_i, x_j)$, $\Omega_{ij}$ is the information matrix of $r_{ij}$.

Then, loop closure detection is carried out by registering adjacent key-frames' local map. And the corresponding error is added into (18). Finally, the total residual error is minimized. The accuracy of the SLAM result is improved, and a globally consistent map is accessed.

$$F(x) = \sum_{<i,j> \in C} r_{ij}^T \Omega_{ij} r_{ij} \quad (18)$$

$$r(\Pi^w, T_j^w) = \Pi^w - d^j R_j^w \vec{n}^j - \left(R_j^w \vec{n}^j\right)^T t_j^w \left(R_j^w \vec{n}^j\right) \quad (19)$$

$$J_{\Pi^w} = I_{3\times3} \quad (20)$$

$$J_{R_j^w} \approx -d^j \left(R_j^w \vec{n}^j\right)^{\wedge} \quad (21)$$

$$J_{t_j^w} = R_j^w \vec{n}^j \left(R_j^w \vec{n}^j\right)^T \quad (22)$$

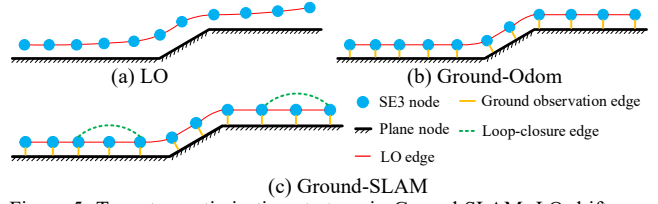

Figure 5. Two-step optimization strategy in Ground-SLAM. LO drifts are compressed by ground observation constraints at the first step, robust loop closure detection is then conducted to access globally consistent map.

## V. EXPERIMENTAL EVALUATION

We evaluate our algorithms on two datasets. The first one is the odometry datasets of KITTI Vision Benchmark. To be specific, we use sequences 00 and 05 that contain abundant planes and almost flat ground. The second one is the HIKVISION dataset which contains two scenarios, the HIKVISION campus and HIKVISION parking lots. A LiDAR with 40 scan-lines configured at 10 Hz, an IMU and a wheel odometer configured at 100 Hz, together with a RTK-GPS running at 10 Hz is mounted on our mapping vehicle. Absolute Trajectory Error (ATE) metric is utilized to compare odometry and SLAM trajectories with RTK-GPS ground-truth trajectory. To be noted, all frames have been utilized to align with ground-truth. We include here LOAM for odometry comparison and LeGO-LOAM for SLAM comparison.

*A. KITTI Vision Benchmark*

All of the experimental statistic analysis results in KITTI sequence 00 and 05 datasets have been listed in Table I, Table II, Table III and Table IV. With these results, we show the influences of our key contributions.

**Sensor-centric sliding map maintenance.** ICP-Odom and ICP-Odom* are the LO approaches that adopt sensor-centric sliding map framework. The only difference between ICP-Odom and ICP-Odom* approaches is that ICP-Odom adopts our proposed observation-based maintenance method, while ICP-Odom* adopts the range-based maintenance method. In this test, range-based maintenance method sets 80 meter as its cut-off range, which is slightly shorter than LiDAR maximal observation range. Both approaches are applied on the KITTI dataset sequences

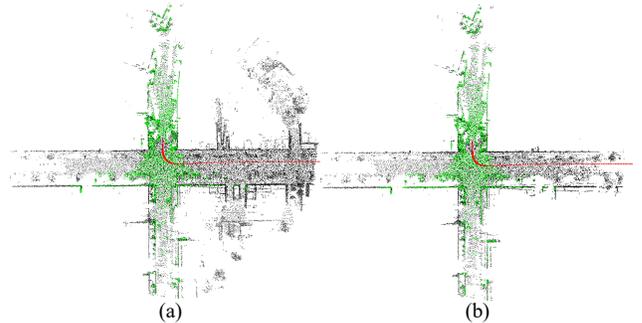

Figure 6. The sliding maps maintained by range-based method (a) and observation-based method (b) on Kitti Sequence 00, snapshot at scan NO.130. The red line is the estimated trajectory. The green point cloud is the currrent scan while the black point cloud is the maintained sliding map.

Table. I Statistics of sliding map's point number and calculate time cost of LO, in KITTI sequence 00 dataset

| Approach | Mean-num | Std-num | Max-nun | Mean-Time[ms] | Std-Time[ms] | Max-Time[ms] |
|---|---|---|---|---|---|---|
| ICP-Odom* | 28348.6 | 5025.3 | 42040 | 133.9 | 31.5 | 611.4 |
| ICP-Odom | 14507.1 | 3179.9 | 25272 | 112.0 | 32.4 | 450.4 |

Table. II Statistics of sliding map's point number and calculate time cost of LO, in KITTI sequence 05 dataset

| Approach | Mean-num | Std-num | Max-nun | Mean-Time[ms] | Std-Time[ms] | Max-Time[ms] |
|---|---|---|---|---|---|---|
| ICP-Odom* | 28216.3 | 5016.3 | 42272 | 139.9 | 33.3 | 639.3 |
| ICP-Odom | 15279.1 | 3400.7 | 27896 | 113.8 | 24.4 | 372.0 |

Table. III RMSE of translation and rotation in KITTI sequence 00 dataset

| Approach | trans-x [m] | trans-y [m] | trans-z [m] | trans [m] | roll [deg] | pitch [deg] | yaw [deg] | rot [deg] |
|---|---|---|---|---|---|---|---|---|
| LOAM | 2.015 | 1.820 | 3.661 | 4.558 | 1.092 | 1.039 | 0.945 | 1.783 |
| ICP-Odom* | 4.374 | 3.881 | 3.158 | 6.646 | 0.925 | 0.811 | 1.399 | 1.863 |
| ICP-Odom | 1.808 | 1.714 | 3.038 | 3.929 | 0.922 | 0.760 | 0.803 | 1.439 |
| Ground-Odom | 1.863 | 1.739 | 0.497 | 2.596 | 0.368 | 0.193 | 0.824 | 0.923 |
| LeGO-LOAM | 0.648 | 0.782 | 1.346 | 1.687 | 0.814 | 0.808 | 0.945 | 1.487 |
| Ground-SLAM | 0.653 | 0.508 | 0.499 | 0.966 | 0.323 | 0.204 | 0.642 | 0.747 |

Table. IV RMSE of translation and rotation in KITTI sequence 05 dataset

| Approach | trans-x [m] | trans-y [m] | trans-z [m] | trans [m] | roll [deg] | pitch [deg] | yaw [deg] | rot [deg] |
|---|---|---|---|---|---|---|---|---|
| LOAM | 1.200 | 1.179 | 2.097 | 2.688 | 0.775 | 0.524 | 0.701 | 1.170 |
| ICP-Odom* | 1.277 | 1.262 | 1.690 | 2.466 | 0.763 | 0.892 | 0.673 | 1.354 |
| ICP-Odom | 1.193 | 1.190 | 1.701 | 2.394 | 0.765 | 0.889 | 0.643 | 1.338 |
| Ground-Odom | 1.187 | 1.186 | 0.632 | 1.793 | 0.438 | 0.451 | 0.646 | 0.902 |
| LeGO-LOAM | 0.681 | 0.565 | 0.860 | 1.234 | 0.516 | 0.533 | 0.592 | 0.949 |
| Ground-SLAM | 0.368 | 0.348 | 0.635 | 0.813 | 0.434 | 0.446 | 0.291 | 0.688 |

Table. V RMSE of translation and rotation in HIKVISION parking lots dataset

| Approach | trans-x [m] | trans-y [m] | trans-z [m] | trans [m] | roll [deg] | pitch [deg] | yaw [deg] | rot [deg] |
|---|---|---|---|---|---|---|---|---|
| LOAM | 0.077 | 0.041 | 0.131 | 0.158 | 0.254 | 0.952 | 1.089 | 1.469 |
| LeGO-LOAM | 0.229 | 0.204 | 0.169 | 0.349 | 0.409 | 0.803 | 1.325 | 1.602 |
| Ground-SLAM | 0.110 | 0.075 | 0.042 | 0.143 | 0.289 | 0.924 | 1.092 | 1.459 |

Table. VI RMSE of translation and rotation in HIKVISION campus dataset

| Approach | trans-x [m] | trans-y [m] | trans-z [m] | trans [m] | roll [deg] | pitch [deg] | yaw [deg] | rot [deg] |
|---|---|---|---|---|---|---|---|---|
| LOAM | 0.357 | 0.332 | 0.210 | 0.531 | 0.402 | 0.444 | 0.464 | 0.758 |
| LeGO-LOAM | 0.577 | 0.720 | 0.348 | 0.986 | 0.467 | 0.513 | 0.676 | 0.969 |
| Ground-SLAM | 0.115 | 0.125 | 0.173 | 0.243 | 0.282 | 0.418 | 0.406 | 0.647 |

00 and 05. Table I and Table II present the statistical result of the sliding map's point number and the calculation time of LO, considering all frames. These results indicate that the observation-based maintenance method efficiently filters out redundant points of the sliding map. Compared with ICP-Odom*, ICP-Odom requires less than 55% points and save more than 16% computational time, for the entire LO process. The improvement brought by observation-based method can be intuitively found in Fig. 6. Meanwhile, higher odometry accuracy is obtained by the ICP-Odom, especially in the sequences 00, as shown in Table III and Table IV.

**Ground observation constraints.** Ground planes are leveraged to compress the pose drift of LO, especially the vertical drift caused by the LiDAR measurement bias. The comparison results between ICP-Odom and Ground-Odom, as shown in Table III and Table IV, further illustrate the influence of ground observation constraints. Generally, Ground-Odom achieves 2.596 m and 0.923 deg of translational and rotational errors, compared to 3.929 m and 1.439 deg of ICP-Odom in KITTI 00 sequence, and achieves 1.793 m and 0.902 deg of translational and rotational errors, compared to 2.394 m and 1.338 deg of ICP-Odom in KITTI 05 sequence. Moreover, the comparison of $z$-translational errors between ICP-Odom and Ground-Odom further demonstrates the ability to compress vertical drifts of ground observation constraints. Roll and pitch errors also decrease as expected since ground observation constraints can also provide constraints for these two angles.

**Loop closure constraints.** Since LO inevitably drifts, loop closure plays important roles in the SLAM system to improve the accuracy of the estimated trajectory. Therefore, although loop closure detection and its constraints are not

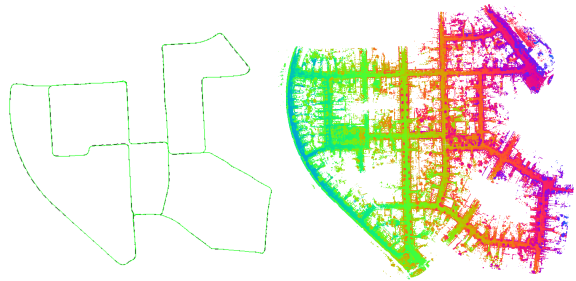
(a) KITTI sequence 00

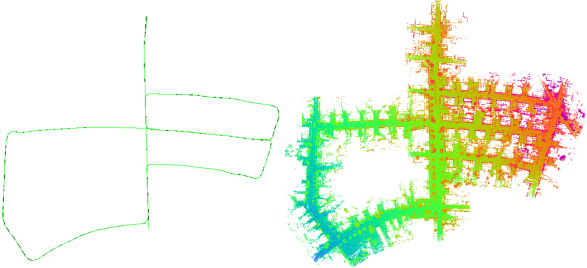
(b) KITTI sequence 05

Figure 7. Trajectories and resulting point clouds of Ground-SLAM in KITTI 00 and 05 sequence. The dashed black trajectory corresponds to the ground-truth and the green to Ground-SLAM.

vital contributions in this paper, Ground-SLAM's results are discussed here. After adding loop closure constraints to Ground-Odom, translational and rotational errors of Ground-SLAM are decreased significantly as expected. Specifically, translational and rotational errors have been compressed to 0.966 m and 0.747 deg in KITTI 00 sequence, 0.813 m and 0.688 deg in KITTI 05 sequence, respectively. Although LeGO-LOAM is a state-of-the-art LiDAR SLAM algorithm that also utilizes loop closure constraints and ground features, Ground-SLAM achieves comparable or better pose estimation accuracy in this test.

### B. HIKVISION Dataset

Further experiments have been conducted in HIKVISION datasets. In this datasets, we only compare Ground-SLAM with LOAM and LeGO-LOAM since the individual influences of each contribution have been demonstrated in the public KITTI dataset.

Experimental results in the HIKVISION parking lots dataset are shown in Table V and Table VI. Generally, Ground-SLAM performs on par with LOAM and LeGO-LOAM. And Ground-SLAM often achieves better results in terms of z-translational error as expected.

### C. Underground Multi-Floor Parking Lots

As indoor multi-floor environments are the main target of this paper, further experiments in underground multi-floor parking lots have also been conducted. The ground-truth of the trajectories are not available in this environment at present. Thus we compare our mapping point clouds with LOAM and LeGO-LOAM's resulting point clouds instead of evaluating trajectories. The mapping points of Ground-SLAM, LOAM and LeGO-LOAM are shown in Fig. 10. From Fig. 10, the contour of different objects of Ground-SLAM results is sharp and clear, such as cars and walls. The grounds of the B2 floor

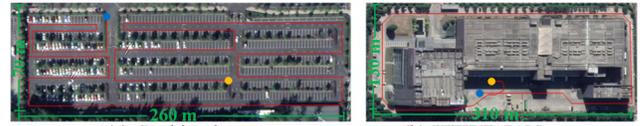
(a) Parking lots      (b) HIKVISION campus

Figure 8. Satellite map of the open flat parking lot and HIKVISION campus. Blue and orange circles represent starting point and ending point respectively, red lines are data collecting routes.

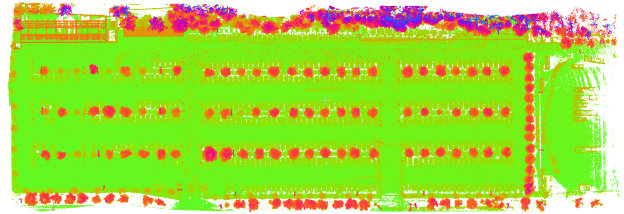
(a) HIKVISION parking lots

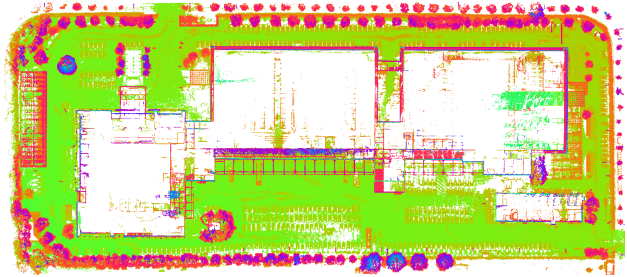
(b) HIKVISION campus

Figure 9. Resulting point clouds of Ground-SLAM in the HIKVISION dataset, color encodes height.

and the B3 floor are quite flat and thin. Besides, the B2 floor and the B3 floor are parallel to each other. These results typically represent mapping results with high accuracy. By contrast, the details of LOAM and LeGO-LOAM's results are blurred.

## VI. CONCLUSION

In this paper, we propose a ground-constrained LiDAR SLAM algorithm named Ground-SLAM. A novel observation-based maintenance method for the sensor-centric sliding map is proposed. And ground observation constraints are utilized to improve the accuracy of SLAM results. Ground-SLAM is initially designed for the structured multi-floor environment, where the global pose observation is often absent, and the vertical drift of LO is difficult to be diminished. However, Ground-SLAM can work properly when running on outdoor scenarios as long as their grounds are roughly flat. As our experimental evaluation illustrates, Ground-SLAM achieves comparable or better pose estimation results than the state-of-the-art SLAM/LO algorithms in indoor multi-floors and even some outdoor environments.

The major drawback of Ground-SLAM is the adaptability of outdoor scenes. Theoretically, Ground-SLAM might fail in specific environments where the slope of grounds varies gradually, leading to improper ground plane correspondence. We furthermore plan to integrate semantic information to improve the accuracy and robustness of LiDAR SLAM results.

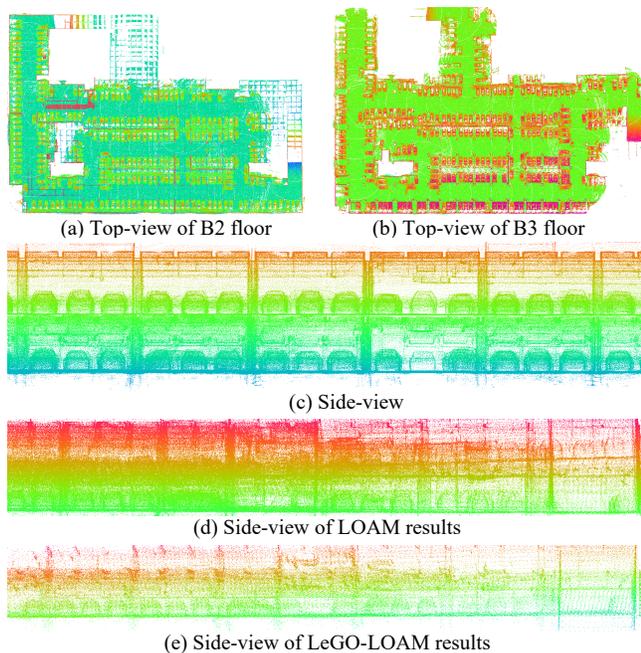

(a) Top-view of B2 floor  (b) Top-view of B3 floor

(c) Side-view

(d) Side-view of LOAM results

(e) Side-view of LeGO-LOAM results

Figure 10. (a)(b)(c) Mapping points of Ground-SLAM in underground multi-floor parking lots, color encodes height. (d) Side-view of LOAM mapping results, two floor is non-parallel. (e) Side-view of LeGO-LOAM mapping result, two floor is partially overlapped.

## VII. Acknowledgement

This work has been supported by National Key R&D Program of China (Grant No.2020AAA010400X).